\documentclass{article}
\usepackage{microtype}
\usepackage{graphicx}
\usepackage{booktabs} 
\usepackage[table]{xcolor}
\usepackage{colortbl}
\usepackage[preprint]{icml2026}
\newcommand{\best}[1]{\cellcolor{green!18}\textbf{#1}}
\newcommand{\second}[1]{\cellcolor{yellow!18}#1}
\usepackage{hyperref}

\usepackage{amsmath}
\usepackage{amssymb}
\usepackage{mathtools}
\usepackage{amsthm}

\theoremstyle{plain}

\theoremstyle{definition}

\theoremstyle{remark}

\icmltitlerunning{DouC: Dual-Branch CLIP for Training-Free Open-Vocabulary Segmentation}

\begin{document}

\twocolumn[
  \icmltitle{DouC: Dual-Branch CLIP for Training-Free Open-Vocabulary Segmentation}



  \icmlsetsymbol{equal}{*}

\begin{icmlauthorlist}
  \icmlauthor{Mohamad Zamini}{equal,uwyo}
  \icmlauthor{Diksha Shukla}{equal,uwyo}
\end{icmlauthorlist}

\icmlaffiliation{uwyo}{Department of Electrical Engineering and Computer Science, University of Wyoming, Laramie, WY, USA}

\icmlcorrespondingauthor{Mohamad Zamini}{mzamini@uwyo.edu}
\icmlcorrespondingauthor{Diksha Shukla}{dshukla@uwyo.edu}


  \vskip 0.3in
]



\printAffiliationsAndNotice{\icmlEqualContribution}

\begin{abstract}
Open-vocabulary semantic segmentation requires assigning pixel-level semantic labels while supporting an open and unrestricted set of categories. Training-free CLIP-based approaches preserve strong zero-shot generalization but typically rely on a single inference mechanism, limiting their ability to jointly address unreliable local tokens and insufficient spatial coherence. We propose \textbf{DouC}, a training-free dual-branch CLIP framework that decomposes dense prediction into two complementary components. OG-CLIP improves patch-level reliability via lightweight, inference-time token gating, while FADE-CLIP injects external structural priors through proxy attention guided by frozen vision foundation models. The two branches are fused at the logit level, enabling local token reliability and structure-aware patch interactions to jointly influence final predictions, with optional instance-aware correction applied as post-processing. DouC introduces no additional learnable parameters, requires no retraining, and preserves CLIP’s zero-shot generalization. Extensive experiments across eight benchmarks and multiple CLIP backbones demonstrate that DouC consistently outperforms prior training-free methods and scales favorably with model capacity.
\end{abstract}
\section{Introduction}

Open-vocabulary semantic segmentation (OVSS) aims to assign pixel-level semantic labels while supporting an open and potentially unbounded set of categories \cite{bai2025self} \cite{lan2024proxyclip}. Unlike conventional semantic segmentation, OVSS must generalize beyond a fixed label space, making it substantially more challenging. Vision–language models (VLMs), most notably CLIP \cite{radford2021learning}, have demonstrated strong zero-shot generalization by learning aligned image–text representations from large-scale web data \cite{li2023blip}. This capability has made CLIP a natural foundation for OVSS. However, CLIP is pretrained for global image–text alignment rather than dense localization, leading to noisy and spatially inconsistent patch-level representations \cite{lan2024proxyclip,chi2025plug}. Additionally, semantic errors introduced at intermediate layers tend to propagate through the network, ultimately corrupting final segmentation logits \cite{jin2025feature}. 

Building on this insight, exiting CLIP-based OVSS approaches broadly fall into two paradigms. Finetuning-based methods adapt CLIP to dense prediction tasks using labeled data, often achieving strong performance but at the cost of additional training, task-specific supervision, and degraded generalization \cite{cha2023learning,wang2024sam,wysoczanska2024clip}. In contrast, training-free methods preserve CLIP’s pretrained representations and rely on architectural or inference-time modifications to enable dense prediction \cite{lan2024clearclip,wang2024sclip,lan2024proxyclip,zhang2025corrclip}. These methods are attractive due to their efficiency, modularity, and robustness across datasets and class vocabularies.


Most existing training-free approaches focus on modifying CLIP’s attention mechanisms to improve spatial coherence. Some suppress unreliable tokens by filtering attention outliers \cite{jin2025feature,chi2025plug}, while others replace or reweight query–key interactions using self-self or proxy attention \cite{lan2024clearclip,wang2024sclip,lan2024proxyclip}. While these strategies improve intermediate representations, they suffer from two key limitations.

First, outlier suppression alone is insufficient: removing unreliable tokens prevents attention over-activation but does not actively strengthen semantic regions or improve class discrimination. Second, intermediate attention refinement is weakly coupled to final predictions. Enhancing attention maps does not guarantee improved segmentation, since subsequent operations may override or dilute these refinements \cite{chi2025plug}. As a result, improvements at the representation level do not always translate into better outputs.

Separately, feedback-driven approaches have shown that output predictions contain valuable semantic structure that can be leveraged to refine intermediate representations \cite{zhang2025corrclip}. By measuring semantic similarity between patch-level predictions and feeding this information back into attention, these methods improve consistency between internal representations and final outputs. However, such approaches operate on a single branch and remain sensitive to prediction noise, especially in cluttered scenes or under domain shift. These observations motivates us combining complementary inductive biases, rather than relying on a single modification to CLIP.

We propose \textbf{DouC}, a dual-branch, training-free CLIP framework for open-vocabulary semantic segmentation. DouC decomposes dense prediction into two parallel but complementary branches. 
The OG-CLIP branch operates directly on OpenAI CLIP features and incorporates lightweight, inference-time token reliability gating to suppress unreliable patch tokens, improving local feature stability without modifying the backbone \cite{jin2025feature}.
The FADE-CLIP branch augments CLIP with feedback-driven semantic adaptation, leveraging DINO \cite{zhang2022dino} to inject output-derived semantic structure back into intermediate representations \cite{zhang2025corrclip}.

The two branches produce dense logit maps that are fused at the logit level, ensuring that both local reliability and global semantic coherence directly influence the final prediction. Importantly, DouC is training-free, introduces no additional learnable parameters, and preserves CLIP’s original generalization capability. To further enhance spatial consistency, DouC integrates instance-level mask correction using off-the-shelf mask generators \cite{kirillov2023segment,cheng2022masked,kerssies2025your}, enabling post-hoc refinement of object boundaries while leaving the core model representations unchanged.
\section{Related Work}

OVSS extends traditional segmentation by requiring generalization to unseen or arbitrary class labels \cite{bai2025self,lan2024proxyclip}. Recent approaches leverage vision–language alignment for zero-shot semantic understanding at the pixel level \cite{wang2024sclip,lan2024clearclip}.

Several methods finetune CLIP or CLIP-derived models for segmentation tasks, often incorporating pixel-level supervision or auxiliary segmentation heads \cite{wysoczanska2024clip,cha2023learning,wang2024sam}. While effective, finetuning compromises CLIP’s pretrained generalization and requires additional labeled data and compute, limiting scalability.


Training-free methods aim to adapt CLIP to dense prediction without modifying its parameters. Prior works improve localization by modifying attention mechanisms, suppressing outlier tokens, or replacing query–key attention with alternative formulations such as self-self or proxy attention \cite{lan2024clearclip,wang2024sclip,lan2024proxyclip,chi2025plug}. These methods improve spatial coherence but primarily operate on intermediate representations and do not explicitly leverage output-level semantic structure.

More recent approaches explore feedback mechanisms that incorporate output predictions back into intermediate attention, enforcing consistency between internal representations and final outputs \cite{zhang2025corrclip}. While effective, these methods typically operate on a single branch and remain sensitive to noisy predictions in complex scenes.




DouC differs from prior work by explicitly integrating two complementary training-free mechanisms—token reliability gating and feedback-driven semantic adaptation—within a unified dual-branch framework. Rather than intervening in a single attention pathway, DouC fuses local token reliability and global semantic coherence directly at the logit level, enabling robust and scalable open-vocabulary segmentation without retraining.

In practice, training-free OVSS benefits from three complementary priors:
\begin{itemize}
\item \textbf{Token reliability priors} within CLIP (e.g., outlier suppression or reliability gating), which mitigate the influence of spurious or noisy tokens.
\item \textbf{External visual priors} (e.g., DINO features or attention-based saliency), which provide geometry- and structure-aware cues for patch interactions.
\item \textbf{Instance-aware constraints} (e.g., SAM2 masks), which enforce spatial consistency within object-like regions and correct local label noise.
\end{itemize}
DouC unifies these priors in a two-branch architecture with logit-level fusion, allowing each signal to contribute independently while remaining fully training-free.

\section{Preliminaries and Motivation}
\label{sec:prelim_motivation}


Let an image be $I\in\mathbb{R}^{3\times H\times W}$ and the set of query strings be $\{q_j\}_{j=1}^{Q}$, where each query $q_j$ is mapped to a dataset class index via a fixed mapping $\pi(j)\in\{1,\dots,C\}$. We seek dense per-query logits $\mathbf{L}\in\mathbb{R}^{Q\times H\times W}$ and, per-class logits $\mathbf{L}^{cls}\in\mathbb{R}^{C\times H\times W}$.

CLIP uses a Vision Transformer (ViT) image encoder that embeds an image into a sequence of tokens. The image is partitioned into $L$ patches, each embedded into $\mathbb{R}^{v}$, and a learnable class token is prepended:
\begin{equation}
\mathbf{X}_{0}=[\mathbf{x}_{cls},\mathbf{x}_{1},\ldots,\mathbf{x}_{L}]\in\mathbb{R}^{(L+1)\times v}.
\end{equation}
A positional embedding is added, and the sequence is processed by a stack of transformer blocks. Each block contains multi-head self-attention (MHSA) and a feed-forward network (FFN). For a generic block input $\mathbf{X}$, the projected queries/keys/values are
\begin{equation}
\mathbf{Q}=\mathbf{X}W_Q,\quad \mathbf{K}=\mathbf{X}W_K,\quad \mathbf{V}=\mathbf{X}W_V,
\end{equation}
and the attention weights are
\begin{equation}
\mathbf{A}=\mathrm{Softmax}\!\left(\frac{\mathbf{Q}\mathbf{K}^{\top}}{\sqrt{v}}\right),\qquad
\mathrm{Attn}(\mathbf{X})=\mathbf{A}\mathbf{V}.
\end{equation}
Dense patch features are obtained by discarding the CLS token and reshaping the patch tokens back into a 2D grid.

Given each query string $q_j$, we build a set of prompts and encode them with the CLIP text encoder. Let $\mathbf{t}_{j}\in\mathbb{R}^{d}$ be the normalized mean embedding of the prompt ensemble for query $q_j$. Stacking yields
\begin{equation}
\mathbf{T}=[\mathbf{t}_1;\ldots;\mathbf{t}_Q]\in\mathbb{R}^{Q\times d}.
\end{equation}
If $Q\neq C$ (e.g., synonyms per class), we later collapse query logits to class logits using the known mapping $\pi$.

Let $\mathbf{Z}\in\mathbb{R}^{L\times d}$ be the normalized dense patch features in the shared vision--language space. The per-patch query logits are computed by cosine similarity:
\begin{equation}
\mathbf{Y}\in\mathbb{R}^{L\times Q}=\mathbf{Z}\mathbf{T}^{\top}.
\end{equation}
Reshaping $\mathbf{Y}$ to the spatial grid gives logits $\mathbf{L}\in\mathbb{R}^{Q\times h\times w}$ at patch resolution, which are upsampled to $H\times W$ as needed.

\begin{figure}
    \centering
    \includegraphics[width=1\linewidth]{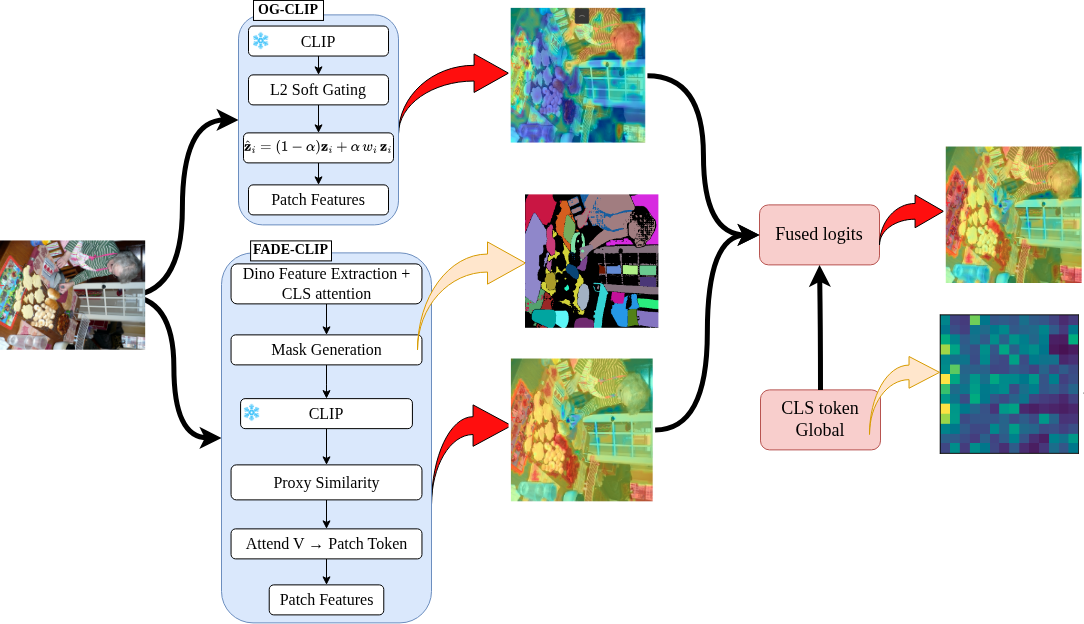}
    \caption{Our Proposed archtiecture}
    \label{fig:placeholder}
\end{figure}
\subsection{Two-Branch Hybrid OVSS}
\label{sec:unified_method}

While CLIP provides strong zero-shot alignment between images and text, its vision encoder is optimized for global recognition and does not explicitly enforce spatial coherence at the patch level. As a result, directly repurposing CLIP for dense prediction leads to two complementary failure modes: (\emph{i}) patch features that are semantically aligned but spatially noisy, and (\emph{ii}) spatially coherent groupings that are weakly aligned with language. To address these limitations in a training-free manner, we design a two-branch architecture that disentangles semantic alignment from spatial structuring and recombines them at the logit level. Given an input image \(I\), we compute two dense query-logit maps:
\begin{itemize}
    \item \textbf{OG-CLIP branch.}  
    This branch preserves the original OpenAI CLIP \cite{radford2021learning} vision--language alignment and produces dense logits
    \(\mathbf{L}_{\mathrm{og}}\in\mathbb{R}^{Q\times H\times W}\) from CLIP patch embeddings. However, dense CLIP features are often dominated by background or low-saliency patches. To mitigate this effect, we introduce a \emph{token purification} mechanism within the CLIP visual transformer that softly down-weights unreliable patch tokens based on their feature magnitude. The purification is applied in a residual, layer-selective manner and does not modify CLIP's training objective or parameters. This stabilizes dense predictions while retaining CLIP’s strong language alignment.

    \item \textbf{FADE-CLIP branch.}  
    This branch explicitly injects spatial structure into CLIP by replacing the final self-attention aggregation with a \emph{proxy affinity} derived from a frozen external vision foundation model, specifically a DINO ViT \cite{caron2021emerging}. DINO patch features induce an affinity matrix that reconstructs the value vectors of the final CLIP transformer block, yielding spatially coherent dense embeddings prior to projection into the shared vision--language space. When instance masks are available, the proxy affinity is further constrained to operate only within individual instance regions, preventing inter-object feature bleeding. The resulting dense logits
    \(\mathbf{L}_{\mathrm{fade}}\in\mathbb{R}^{Q\times H\times W}\) exhibit strong object-level consistency while remaining fully training-free.
\end{itemize}

The two branches address complementary deficiencies: OG-CLIP emphasizes semantic alignment but lacks spatial regularity, whereas FADE-CLIP enforces spatial coherence but relies on external structure. We therefore fuse their predictions linearly:
\begin{equation}
\mathbf{L}_{\mathrm{fused}}
=
\alpha_{\mathrm{fade}}\mathbf{L}_{\mathrm{fade}}
+
\alpha_{\mathrm{og}}\mathbf{L}_{\mathrm{og}},
\label{eq:fusion_basic}
\end{equation}
where \(\alpha_{\mathrm{fade}}\) and \(\alpha_{\mathrm{og}}\) balance spatial consistency against language fidelity.

Finally, to recover global semantic context that may be diluted in dense prediction, we inject a global CLS prior computed from the OG-CLIP image embedding:
\begin{equation}
\mathbf{L}_{\mathrm{fused}}
\leftarrow
\mathbf{L}_{\mathrm{fused}}
+
\lambda_{\mathrm{cls}}\;\mathbf{l}^{\mathrm{cls}}[:, None, None],
\label{eq:cls_prior}
\end{equation}
where \(\mathbf{l}^{\mathrm{cls}}\in\mathbb{R}^{Q}\) denotes query logits obtained from the global image representation. Final predictions are obtained by applying a softmax over queries, collapsing synonymous queries to classes when \(Q\neq C\), followed by an \(\arg\max\) over classes.

\subsection{FADE-CLIP Branch: DINO-Guided Proxy Attention}
\label{sec:fsa_branch}

The FADE-CLIP branch introduces a training-free mechanism that injects dense external visual structure into the CLIP vision transformer by \emph{reconstructing patch-level features in the final transformer block} using DINO-guided proxy attention. The method preserves the original CLIP architecture and parameters, modifying only the inference-time behavior of the last attention block.

\paragraph{External dense features from DINO.}
Given an input image $\tilde{I}$, we extract last-layer patch features from a frozen DINO vision transformer and reshape them into a spatial grid:
\begin{equation}
\mathbf{E}\in\mathbb{R}^{H_f W_f \times C_f}.
\end{equation}
These features, derived from the final DINO attention representations, encode fine-grained semantic and spatial relationships between image regions and serve as an external structural prior for guiding CLIP patch interactions.

\paragraph{DINO-guided proxy affinity.}
We construct a patch-to-patch affinity matrix using normalized dot products between DINO features:
\begin{equation}
s_{ij}=\left\langle \frac{\mathbf{e}_i}{\|\mathbf{e}_i\|_2},\frac{\mathbf{e}_j}{\|\mathbf{e}_j\|_2}\right\rangle,
\end{equation}
where $\mathbf{e}_i$ denotes the DINO feature at spatial location $i$. The affinities are scaled by a temperature parameter $\tau$ and normalized via a row-wise softmax to obtain a proxy attention matrix:
\begin{equation}
\mathbf{A}^{\mathrm{proxy}}=\mathrm{Softmax}(\tau\,\mathbf{S}).
\label{eq:proxy_attn}
\end{equation}

\paragraph{Instance-aware affinity masking.}
When instance segmentation masks are available, the proxy affinities are gated to permit attention only between patches belonging to the same instance. This constraint is applied directly to the affinity matrix prior to normalization, preventing inter-instance feature leakage while preserving the underlying attention formulation.

\paragraph{Value reconstruction in CLIP space.}
Instead of modifying CLIP queries or keys, we extract the value projections from the \emph{final CLIP attention block} and interpolate them to the DINO feature resolution. We then \emph{replace the original self-attention aggregation} with a DINO-guided reconstruction:
\begin{equation}
\mathbf{V}^{\mathrm{proxy}}=\mathbf{A}^{\mathrm{proxy}}\mathbf{V},
\end{equation}
where $\mathbf{V}$ denotes the CLIP value embeddings. This operation injects DINO-driven spatial structure while preserving the CLIP embedding space and its language alignment.

\paragraph{Token reassembly and projection.}
The reconstructed patch tokens are concatenated with the original CLIP class token and passed through the standard CLIP post-processing head (LayerNorm followed by a linear projection). The resulting dense visual embeddings are used to compute open-vocabulary segmentation logits via dot products with text embeddings.

By applying DINO-guided proxy attention exclusively to the value stream of the final CLIP transformer block, FADE-CLIP enhances spatial coherence and object-level consistency while retaining the semantic alignment and zero-shot generalization properties of CLIP.




\subsection{OG-CLIP Branch: Dense Similarity with Outlier-Gated CLIP}
\label{sec:ogclip_branch}

The OG-CLIP branch leverages dense patch-level visual features and open-vocabulary similarity scores derived from the original OpenAI CLIP model. To mitigate the influence of background-dominated or unreliable patch tokens, we introduce a lightweight, training-free \emph{outlier-gating module} that operates directly within the CLIP visual backbone at intermediate transformer layers. In the final model, this module is instantiated using an $\ell_2$-based token reliability criterion, while alternative strategies are evaluated in ablation for completeness.

\paragraph{$\ell_2$-based token reliability.}
Let $\mathbf{z}_i \in \mathbb{R}^{C}$ denote the feature vector of patch token $i$ extracted from an intermediate layer of the CLIP vision transformer. We quantify token reliability using its $\ell_2$ norm,
\begin{equation}
s_i = \|\mathbf{z}_i\|_2,
\end{equation}
which serves as a simple and stable proxy for feature saliency. Under CLIP’s contrastive training objective, feature magnitude correlates with semantic activation strength, causing object-centric patches to exhibit stronger responses than background regions.

\begin{table*}[t]
\centering
\footnotesize
\setlength{\tabcolsep}{4.5pt}
\caption{Comparison of training-free open-vocabulary semantic segmentation methods across different OpenAI CLIP backbones.
All results are mIoU (\%). CorrCLIP and our architecture uses SAM2 \cite{caron2021emerging} for benchmarking.}
\begin{tabular}{l l c c c c c c c c c }
\toprule
\textbf{Model} & \textbf{Venue}  &
\multicolumn{3}{c}{\textbf{with background}} &
\multicolumn{5}{c}{\textbf{without background}} &
\textbf{Avg} \\
\cmidrule(lr){3-5} \cmidrule(lr){6-10}
 &  & 
VOC21 & C60 & Object &
VOC20 & C59 & Stuff & ADE & City &  \\
\midrule
\multicolumn{11}{l}{\textbf{ViT-B Backbone}} \\
\midrule

SCLIP \cite{wang2024sclip} & ECCV'24
& 59.1&30.4&30.5&80.4&34.2&22.4&16.1&32.3&38.2 \\
NACLIP \cite{hajimiri2025pay}  & WACV'25
& 58.9&32.2&33.2&79.7&35.2&23.3&17.4&35.5 &39.5\\
SFP \cite{jin2025feature} & ICCV'25
& 63.9 & 37.2&37.9&84.5&39.9&26.4&20.8&41.1&44\\

FSA \cite{chi2025plug} & ICCV'25 
& 63.7&36.1&38.0&82.3&39.9&27.0&20.5&38.8 &43.3\\

Trident \cite{shi2025harnessing} & ICCV'25 
& 67.1&38.6&41.1&84.5&42.2&28.3&21.9&42.9 &45.8\\

DIH-CLIP \cite{duan2025dih} & ICCV'25 
& 64.2&36&37.4&84.9&39.7&26.7&19.6&40.2&43.6\\

CDAM \cite{kangclass} & ICLR'25
& 58.7&30.6&35.2&-&-&24.8&17.2&23.7 &-\\
CASS \cite{kim2025distilling} & CVPR'25
& 65.8&36.7&37.8&87.8&40.2&26.7&20.4&39.4 &44.4\\
CorrCLIP \cite{zhang2025corrclip}& ICCV'25
& \second{72} & \second{41.1} & \second{40.7}
& \second{87.8} & \best{46.4} & \second{30.5} & \second{23.4} & \second{42.0} &\second{48}\\

Ours & --
& \best{72.2} & \best{41.2} & \best{43.8}
& \best{88} & \second{45.2} & \best{31.1} & \best{23.7} & \best{45.3}&\best{48.8}\\
\midrule
\multicolumn{10}{c}{\textbf{ViT-L Backbone}} \\
\midrule
ProxyCLIP \cite{lan2024proxyclip} & ECCV'24&60.6&34.5&39.2&83.2&37.7&25.6&22.6&40.1&42.9\\
FreeDA \cite{barsellotti2024training}  & ECCV'24
& 55.4&38.3&37.4&87.9&43.5&28.8&23.2&36.7&43.9\\
SC-CLIP \cite{bai2025self} &TIP'25&65.0&36.9&40.5&88.3&40.6&26.9&21.7&41.3&45.2\\
FSA \cite{chi2025plug} & ICCV'25&61.8&34.9&40.2&84.1&38.1&25.9&22.9&41.2&43.6\\
CLIPer \cite{sun2025cliper} & ICCV'25
& 69.8 & 38.0 & 43.3 & 90.0 & 43.6 & 28.7 & 24.4& 41.6  & 47.3\\
ResCLIP \cite{yang2025resclip} & CVPR'25
& 54.1 & 30.9 & 32.5 & 85.5 & 34.5 & 23.4 & 18.2 & 33.7  &  39.1\\
CorrCLIP \cite{zhang2025corrclip}& ICCV'25
& \second{71.2} & \second{41.6} & \second{45.7}
& \second{90.7} & \second{46.1} & \second{30.8} & \second{26.7} & \second{46.3} &\second{49.9}\\

Ours & --
& \best{73.1} & \best{42.8} & \best{47.2}
& \best{90.8} & \best{47.7} & \best{33.1} & \best{27.8} & \best{47.2} &\best{51.2}\\
\midrule
\multicolumn{10}{c}{\textbf{ViT-H Backbone}} \\
\midrule
ProxyCLIP \cite{lan2024proxyclip} & ECCV'24&65.0&35.4&38.6&83.3&39.6&26.8&24.2&42&44.4\\
FSA \cite{chi2025plug} & ICCV'25&67.9&36.3&40.2&85.7&40.5&27.3&24.5&43.6&45.8\\

Trident \cite{shi2025harnessing} & ICCV'25 
&70.8&40.1&42.2&88.7&44.3&28.6&26.7&47.6&48.6\\
CorrCLIP \cite{zhang2025corrclip}& ICCV'25
& \best{75.5} & \second{41.7} & \best{48.1}
& \second{91.4} & \second{46.9} & \second{32.1} & \second{28.1} & \second{47.3} &\second{51.4}\\

Ours & --
& \second{74.3} & \best{43.3} & \second{47.8}
& \best{92.3} & \best{48.6} & \best{33.4} & \best{29.8} & \best{50.1} &\best{52.4}\\

\bottomrule
\end{tabular}
\label{tab:ovss_backbone_comparison}
\end{table*}
\paragraph{Soft outlier gating.}
The raw scores $\{s_i\}$ are mapped to soft gating weights $w_i \in (0,1)$ via a normalized, temperature-controlled transformation. Rather than hard pruning, we apply a residual-style attenuation to each patch token:
\begin{equation}
\hat{\mathbf{z}}_i
= (1-\alpha)\,\mathbf{z}_i + \alpha\, w_i\, \mathbf{z}_i,
\label{eq:ogclip_gate}
\end{equation}
where $\alpha \in [0,1]$ controls the strength of suppression. This formulation preserves the original feature direction while adaptively modulating token magnitude, enabling gradual down-weighting of unreliable patches without disrupting the transformer structure or CLIP’s vision--language alignment. The gated features are then used directly for dense similarity computation with text embeddings.

\section{Experiments}
\subsection{Implementation Details}

We evaluate our approach against a diverse set of strong training-free open-vocabulary semantic segmentation baselines. Specifically, we consider CLIP \cite{radford2021learning} with ViT-B/16 and ViT-L/14 backbones, as well as OpenCLIP \cite{cherti2023reproducible} with a ViT-H/14 backbone. For fair comparison, DouC-CLIP is evaluated across multiple backbone scales to assess robustness under varying model capacities.

For external affinity induction, we employ DINOv1 \cite{caron2021emerging}, DINOv2 \cite{oquab2023dinov2}, and MAE \cite{he2022masked}, using DINOv1 as the default setting unless otherwise specified. For instance-guided correction, we consider Mask2Former \cite{cheng2022masked}, EoMT \cite{kerssies2025your}, and SAM2 \cite{ravi2024sam}, and adopt SAM2 as the baseline instance guidance mechanism in all main experiments.

To further examine the impact of the underlying vision--language backbone, we evaluate our method across multiple CLIP-style pretraining variants, including DFN \cite{fang2023data}, MetaCLIP \cite{xu2023demystifying}, LAION-CLIP \cite{cherti2023reproducible}, and the original CLIP \cite{radford2021learning}.
\subsection{Comparison with State-of-the-Art Methods}
\label{sec:comparison}
Table \ref{tab:ovss_backbone_comparison} compares our method against representative training-free OVSS approaches across multiple benchmarks and CLIP backbone scales. All methods are evaluated under a unified, training-free protocol with identical visual backbones and SAM2-based instance masks when applicable. We include recent and competitive baselines such as SFP, FSA, Trident, DIH-CLIP, and CorrCLIP.

Across all backbone scales, our approach achieves the highest average mIoU, demonstrating consistent gains over prior work. The improvements are particularly pronounced on large-scale and object-centric benchmarks such as \emph{Object}, \emph{Stuff}, \emph{ADE20K}, and \emph{Cityscapes}, which require both fine-grained spatial coherence and robust semantic alignment. These results validate the complementary roles of OG-CLIP token reliability modeling and FADE-CLIP’s DINO-guided proxy attention in stabilizing dense predictions under cluttered and high-variation scenes.


These results highlight that explicitly decoupling semantic reliability (OG-CLIP) from spatial structure injection (FADE-CLIP), and fusing them at the logit level, provides a more robust and scalable solution for training-free open-vocabulary semantic segmentation than prior single-branch or early-injection designs.

\section{Ablation Study}
\vspace{-6pt}
\subsection{Effect of Individual Branches}
\label{sec:branch_effect}

By default, the fusion weights $\alpha_{\text{OG}}$ and $\alpha_{\text{FADE}}$ are set to $0.5$. To isolate the contribution of each component in our hybrid architecture, we evaluate the OG-CLIP and FADE-CLIP branches independently under identical settings, including the same backbone, text prompts, and inference protocol. Table \ref{tab:branch_effect} reports mIoU results across representative benchmarks.

\begin{table}[htbp]
\centering
\footnotesize
\setlength{\tabcolsep}{6pt}
\caption{Effect of individual branches evaluated independently.
OG and FADE denote pure branches.
OG$\rightarrow$FADE and FADE$\rightarrow$OG indicate asymmetric fusion
with 75\% contribution from the source branch and 25\% from the target branch.
All results are mIoU (\%).}

\label{tab:branch_effect}
\begin{tabular}{lcccc}
\toprule
\rotatebox{0}{\textbf{Dataset}} &
\rotatebox{0}{\textbf{OG}} &
\rotatebox{0}{\textbf{FADE}} &
\rotatebox{0}{\textbf{OG$\rightarrow$FADE}} &
\rotatebox{0}{\textbf{FADE$\rightarrow$OG}} \\
\midrule
ADE20K     & 21.4 & 22.8 & \textbf{23.0} & 22.9 \\
Cityscapes & \textbf{44.9} & 43.3 & 43.6 & 38.5 \\
Object     & 39.9 & 41.6 & \textbf{43.1} & 43.0 \\
Stuff      & 28.9 & 30.1 & \textbf{30.7} & 30.4 \\
C59        & 44.4 & 43.2 & \textbf{45.1} & 43.6 \\
C60        & 39.0 & 39.5 & \textbf{40.8} & 40.1 \\
VOC20      & 86.2 & 86.2 & \textbf{87.4} & 86.7 \\
VOC21      & 70.4 & 68.1 & \textbf{70.9} & 70.4 \\
\bottomrule
\end{tabular}
\end{table}


The OG-CLIP branch, which relies on dense CLIP patch embeddings with token reliability gating, performs strongly on datasets emphasizing global object presence and clean object boundaries. In particular, OG-CLIP achieves higher performance on \emph{Cityscapes}, \emph{VOC21}, and \emph{C59}, where CLIP’s global semantic alignment and robust category recognition play a dominant role. These results indicate that token-level outlier gating effectively stabilizes dense predictions while preserving CLIP’s zero-shot recognition strength.

The FADE-CLIP branch consistently outperforms OG-CLIP on datasets that require fine-grained spatial reasoning and object-level consistency. Notable improvements are observed on \emph{ADE20K}, \emph{Object}, \emph{Stuff}, and \emph{C60}, reflecting the benefit of DINO-guided proxy attention in enforcing structured patch aggregation. These gains highlight the importance of external spatial priors for handling cluttered scenes and high intra-class variation.

The asymmetric fusion results further clarify the complementary roles of the two branches. OG$\rightarrow$FADE consistently improves upon the pure OG-CLIP branch across most benchmarks, indicating that even a modest FADE contribution provides effective spatial regularization. In contrast, FADE$\rightarrow$OG yields smaller or inconsistent gains relative to pure FADE-CLIP, suggesting that FADE’s performance is driven primarily by its proxy-based aggregation and is less dependent on semantic correction from OG-CLIP. This asymmetric behavior underscores the complementary nature of the two mechanisms. This shows neither branch dominates across all benchmarks: OG-CLIP favors global semantic alignment, while FADE-CLIP excels at spatial regularization. This complementarity directly motivates our dual-branch design, which allows both mechanisms to operate independently and be fused at the logit level, achieving consistently superior performance across diverse datasets.
\subsection{Impact of External Vision Foundation Models}
\label{sec:vfm_effect}

FADE-CLIP relies on an external vision foundation model (VFM) to induce proxy affinities that guide patch-level aggregation in the final CLIP transformer block. To study the role of this external structure, we evaluate FADE-CLIP using different VFMs, including DINO, DINOv2, MAE, as well as a \emph{No-VFM} variant where the proxy attention mechanism is disabled in Table \ref{tab:vfm_effect}. 

\begin{table}[htbp]
\centering
\footnotesize
\setlength{\tabcolsep}{6pt}
\caption{Impact of different external vision foundation models in FADE-CLIP.
All results are mIoU (\%).}
\label{tab:vfm_effect}
\begin{tabular}{lcccc}
\toprule
\textbf{Dataset} & \textbf{DINO} & \textbf{DINOv2} & \textbf{MAE} & \textbf{No VFM} \\
\midrule
ADE20K   & \textbf{23.7} & 23.2 & 23.1 & 7.6 \\
Cityscapes & \textbf{45.3} & 42.7 & 44.0 & 33.6 \\
Object   & \textbf{43.8} & 40.7 & 40.9 & 12.8 \\
Stuff    & \textbf{31.1} & 30.5 & 29.9 & 11.8 \\
C59      & \textbf{45.2} & 44.7 & 45.0 & 32.7 \\
C60      & \textbf{41.2} & 39.9 & 40.3 & 22.3 \\
VOC20    & \textbf{88.0} & 87.3 & 87.6 & 82.7 \\
VOC21    & \textbf{72.2} & 69.6 & 71.5 & 54.1 \\
\bottomrule
\end{tabular}
\end{table}

Removing the external VFM leads to a substantial performance degradation across all benchmarks, with particularly severe drops on dense and cluttered datasets such as \emph{ADE20K}, \emph{Object}, and \emph{Stuff}. This confirms that CLIP’s native self-attention alone is insufficient to recover reliable spatial structure for dense prediction, and highlights the necessity of external guidance for patch-level aggregation.
\begin{figure*}
    \centering
    \includegraphics[width=0.9\linewidth]{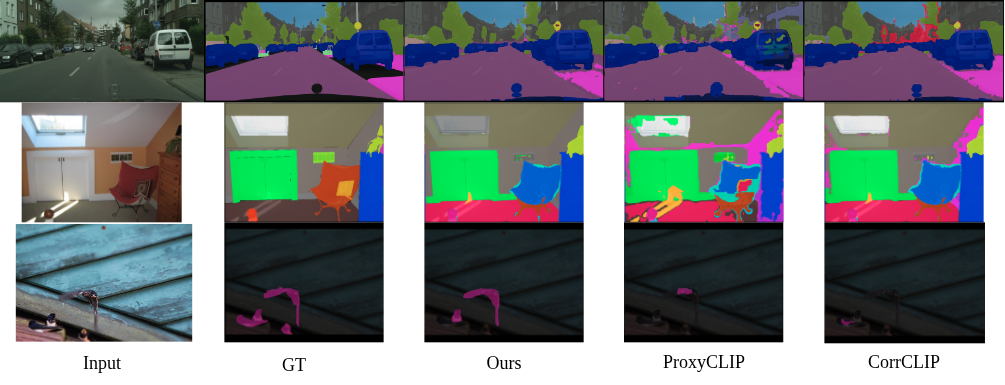}
    \caption{Qualitative comparisons on multiple benchmarks. Rows correspond (top to bottom) to Cityscapes, ADE20K, and COCO-Object. Our method produces more coherent regions and cleaner object boundaries across diverse scenes.}
    \label{fig:qualitative}
\end{figure*}
All three VFMs yield strong and consistent improvements over the No-VFM baseline, demonstrating that FADE-CLIP is not tied to a specific external model. DINO generally performs best on dense and object-centric benchmarks, reflecting its strong patch-level discriminability. MAE achieves competitive results, particularly on category-heavy datasets such as \emph{C59} and \emph{C60}, while DINOv2 exhibits notable gains on \emph{VOC21}, suggesting improved global context modeling. These differences indicate that FADE-CLIP can flexibly exploit complementary inductive biases from different VFMs.

Despite architectural differences among external VFMs, FADE-CLIP consistently benefits from external spatial structure, underscoring the robustness of the proposed proxy attention mechanism. This flexibility enables FADE-CLIP to leverage a wide range of pretrained vision models without retraining, while maintaining strong performance across diverse open-vocabulary segmentation benchmarks.

\subsection{Effect of Instance Mask Generators}
\label{sec:mask_effect}

Instance-level correction plays a complementary role in our framework by enforcing region consistency after dense prediction. To evaluate its impact, we compare different mask generation strategies, including a generic \emph{Mask}-based baseline, SAM, and EoMT, across multiple CLIP backbone scales. All results are reported in mIoU (\%) and use identical backbones, prompts, and inference settings.

Across all backbone scales and datasets, incorporating instance masks consistently improves segmentation performance over the baseline mask strategy. The gains are most pronounced on object-centric and cluttered benchmarks such as \emph{Object}, \emph{Stuff}, and \emph{Cityscapes}, where enforcing region-level consistency reduces boundary noise and inter-instance confusion.

Among the evaluated strategies, EoMT consistently achieves the strongest performance, followed by SAM, with both significantly outperforming the generic mask baseline. EoMT yields particularly large gains on datasets with complex object layouts and high intra-class variation, suggesting that higher-quality and more complete instance proposals better complement FADE-CLIP’s proxy attention and OG-CLIP’s semantic alignment.

The benefit of stronger instance masks increases with backbone capacity. As the CLIP backbone scales from ViT-B to ViT-H, the performance gap between EoMT and weaker mask generators widens, indicating that high-capacity visual encoders are better able to exploit precise instance-level structure during map correction.

\subsection{Qualitative Results}
\label{sec:qualitative}

Figure \ref{fig:qualitative} presents qualitative comparisons on representative images from
\textit{Cityscapes}, \textit{COCO-Object}, and \textit{ADE20K}, illustrating the visual behavior of our method under diverse scene layouts and object scales.
Across all datasets, our approach produces segmentation maps that are both spatially coherent and semantically consistent, complementing the quantitative gains reported in Section \ref{sec:comparison}.
\begin{table}[htbp]
\centering
\footnotesize
\setlength{\tabcolsep}{4pt}
\caption{Impact of different instance mask generators across CLIP backbones.
All results are mIoU (\%). Column headers denote benchmarks and are rotated for compactness. M2F denotes Mask2Former.}
\label{tab:mask_effect}
\begin{tabular}{lcccccccc}
\toprule
 \textbf{Mask} &
\rotatebox{90}{ADE} &
\rotatebox{90}{City} &
\rotatebox{90}{Object} &
\rotatebox{90}{Stuff} &
\rotatebox{90}{C59} &
\rotatebox{90}{C60} &
\rotatebox{90}{VOC20} &
\rotatebox{90}{VOC21} \\
\midrule
\multicolumn{9}{c}{\textbf{ViT-B Backbone}} \\

\midrule
NoMask &21.1 & 40.0&37.8 &27.9 &35.6 &33.8 & 83.3 & 61.9 \\
 M2F & 21.8 & 41.1 & 46.5 & 32.2 & 45.5 & 40.9 & 86.7 & 72.7 \\
 SAM  & 23.7 & 45.3 & 43.8 & 31.1 & 45.2 & 41.2 & 88.0 & 72.2 \\
 EoMT & \best{23.4} & \best{46.3} & \best{46.6} & \best{33.0} & \best{47.3} & \best{42.6} & \best{88.9} & \best{73.8} \\
\midrule
\multicolumn{9}{c}{\textbf{ViT-L Backbone}} \\
\midrule

 M2F & 25.2 & 43.0 & 50.1 & 34.6 & 47.4 & 42.6 & 90.2 & 72.9 \\
 SAM  & 27.8 & 47.2 & 47.2 & 33.1 & 47.7 & 42.8 & 90.8 & 73.1 \\
 EoMT & \best{27.7} & \best{48.2} & \best{50.5} & \best{35.2} & \best{49.6} & \best{44.3} & \best{92.0} & \best{74.3} \\
\midrule
\multicolumn{9}{c}{\textbf{ViT-H Backbone}} \\
\midrule

 M2F & 27.2 & 45.4 & 50.8 & 35.3 & 48.7 & 43.4 & 91.4 & 74.2 \\
 SAM  & 29.8 & 50.1& 47.8 & 33.4 & 48.6 & 43.3 & 92.3 & 74.3 \\
 EoMT & \best{29.7} & \best{50.7} & \best{51.4} & \best{36.2} & \best{51.0} & \best{45.1} & \best{93.6} & \best{76.3} \\
\bottomrule
\end{tabular}
\end{table}

On \textit{Cityscapes}, which features structured outdoor scenes with strong geometry and thin objects, our method yields markedly cleaner road boundaries, improved vehicle separation, and reduced background leakage into foreground regions.
Compared to prior training-free methods, object contours (e.g., cars and sidewalks) are sharper and less fragmented, reflecting the effectiveness of FADE-CLIP’s DINO-guided proxy attention in propagating structure-aware affinities across patches.

On \textit{COCO}, which contains cluttered scenes with overlapping objects and high intra-class variation, our method better preserves object integrity and suppresses spurious activations on background textures.
Notably, small and partially occluded objects are more consistently recovered, indicating that the OG-CLIP branch’s outlier-gated token modeling successfully attenuates noisy patch responses while retaining discriminative object cues.

For \textit{ADE20K}, characterized by complex indoor layouts and fine-grained semantic categories, our predictions exhibit improved region consistency and reduced semantic confusion between visually similar classes.
Walls, furniture, and ceiling regions are more uniformly segmented, with fewer isolated misclassified patches.
This behavior highlights the complementary roles of FADE-CLIP in enforcing spatial coherence and OG-CLIP in stabilizing dense similarity matching.
\subsection{Effect of CLIP Pretraining Source}
\label{sec:pretraining_effect}

Table \ref{tab:vitb_pretraining} analyzes the impact of different ViT-B CLIP pretraining sources on the proposed method.
We evaluate four widely used CLIP variants—DFN, LAION, OpenAI, and MetaCLIP—under an identical inference pipeline and hyperparameter setting, isolating the effect of the underlying pretraining data and objective.

\begin{table}[htbp]
\centering
\footnotesize
\setlength{\tabcolsep}{3.6pt}
\caption{Effect of CLIP ViT-B pretraining source on our method.
All values are mIoU (\%). Best and second-best are marked per row.}
\begin{tabular}{l c c c c}
\toprule
\textbf{Dataset} & \textbf{DFN} & \textbf{LAION} & \textbf{OpenAI} & \textbf{MetaCLIP} \\
\midrule
ADE    & \best{28.0} & \second{27.2} & 27.2 & 26.5 \\
City   & \best{49.9} & 47.0 & 45.3 & \second{48.8} \\
Object & \second{45.4} & 41.3 & 43.5 & \best{46.0} \\
Stuff  & \best{34.1} & 32.0 & 31.1 & \second{33.2} \\
C59    & \best{48.8} & 45.3 & 45.2 & \second{48.3} \\
C60    & \best{43.8} & 40.4 & 41.2 & \second{43.6} \\
VOC20  & \best{89.2} & \second{89.2} & 88.0 & 88.8 \\
VOC21  & \second{72.9} & 70.3 & 71.3 & \best{73.5} \\
\midrule
Avg    & \best{51.5} & 49.1 & 50.6 & \second{51.1} \\
\bottomrule
\end{tabular}
\label{tab:vitb_pretraining}
\end{table}
Overall, our method demonstrates strong robustness across pretraining sources, achieving competitive performance in all cases.
Among them, DFN-pretrained CLIP consistently yields the highest average mIoU, achieving the best results on six out of eight datasets, including ADE20K, Cityscapes, COCO-Stuff, and the C59/C60 benchmarks.
This suggests that higher-quality or more diverse pretraining distributions can further amplify the benefits of our inference-time enhancements.


These results indicate that our hybrid OG-CLIP and FADE-CLIP design is largely orthogonal to the choice of CLIP pretraining source.
Rather than overfitting to a particular model variant, the method consistently improves dense prediction quality by stabilizing patch-level representations and injecting external structural priors.
This robustness further underscores the practicality of the proposed approach for deployment with off-the-shelf CLIP models trained on diverse datasets.

\section{Conclusion}
\label{sec:conclusion}

We propose a simple, training-free hybrid framework for open-vocabulary semantic segmentation that combines two complementary CLIP-based inference mechanisms. The method integrates OG-CLIP, which stabilizes dense similarity matching via token-level outlier gating, with FADE-CLIP, which injects external structural priors through DINO-guided proxy attention applied at inference time. Operating entirely on frozen vision–language models and modifying only the final stages of the visual backbone, our approach preserves CLIP’s zero-shot generalization while substantially improving spatial coherence and object consistency. Extensive experiments across multiple benchmarks and CLIP backbone scales demonstrate consistent gains over prior training-free methods, with ablations confirming the complementary and synergistic roles of the two branches and robustness to different external feature models and mask generators. Qualitative results further corroborate these findings, showing sharper boundaries, reduced semantic leakage, and improved region consistency, highlighting the effectiveness of inference-time adaptation through principled reuse and alignment of representations already present in large vision–language models.

\section*{Impact Statement}

This paper presents work whose goal is to advance the field of Machine
Learning. There are many potential societal consequences of our work, none
which we feel must be specifically highlighted here.
\section*{Acknowledgment}
This work was supported in part by the AI-Seed Research Award from the Research and Economic Development Division at the University of Wyoming. 
\nocite{langley00}

\bibliography{example_paper}
\bibliographystyle{icml2026}




\end{document}